# Different representation learning objectives recover distinct latent structures from the same psychometric data

Cong Cao *[1,2], Tassos C. Kyriakides [1,2], Pambos Vrasidas [3]

1. Department of Biostatistics, Yale School of Public Health, Yale University, New Haven
2. Cooperative Studies Program Coordinating Center, VA Connecticut Healthcare System, West Haven, CT, USA
3. Center for the Advancement of Research & Development in Educational Technology (CARDET), Nicosia, Cyprus

Corresponding author: Cong Cao, cong.cao@yale.edu

**Abstract**: Psychometric questionnaires contain rich item-level information, yet it remains unclear whether different representation learning objectives recover the same latent organization. We investigated this question using 757 matched teacher–child pairs from the baseline assessment of the Cyprus ProW preschool trial. Behavioral structure was characterized from child SDQ, ASBI, and CBRS item responses using principal component analysis and clustering, yielding four behavioral phenotypes. A contrastive objective substantially improved teacher–child retrieval relative to PCA-based representations, increasing Top-1 accuracy from 0.13% to 7.27% and Top-10 accuracy from 1.98% to 56.14%. However, contrastive representations preserved behavioral phenotype structure less effectively than PCA-based representations. A multi-task objective jointly optimizing alignment and behavioral prediction partially restored behavioral organization but reduced retrieval performance. These findings indicate that teacher–child correspondence and behavioral phenotypes represent distinct forms of latent organization and demonstrate that the latent structure recovered from linked psychometric data depends on the representation learning objective.

## 1. Introduction

Teachers influence children's social, emotional, and behavioral development in many ways. Research has shown that teacher wellbeing, self-efficacy, burnout, and classroom climate are associated with children's behavioral and academic outcomes (Hamre et al., 2001; Jennings & Greenberg, 2009; Mashburn et al., 2008). Most studies examine these relationships using aggregated questionnaire scores. Individual questionnaire items are combined into a small number of scale scores and then analyzed using correlation, regression, or structural equation models. Although this approach is well established, it assumes that the most informative structure in the data is adequately represented by these summary scores. Previous research suggests that item-level responses often contain meaningful psychological information that is not fully preserved after score aggregation (Reise, 2012; McNeish, 2020).

Recent advances in deep representation learning have provided new ways to learn informative representations directly from high-dimensional observations (LeCun et al., 2015). Rather than relying on predefined scales, these methods learn low-dimensional representations from item-level responses. Contrastive learning has become one of the most successful approaches for learning such representations (Chen et al., 2020; van den Oord et al., 2018). Transformer-based models and contrastive learning have shown strong performance for structured data, and they are increasingly being applied to behavioral and psychometric research (Vaswani et al., 2017; Huang et al., 2020; Somepalli et al., 2021; Dagum, 2019; Dhelim et al., 2022). Although representation learning is increasingly used in behavioral research, prediction-oriented representations are not necessarily optimized for scientific interpretation (Yarkoni & Westfall, 2017). However, an important question remains unanswered: when the same psychometric data are analyzed using different representation learning objectives, do they recover the same underlying latent structure?

Linked teacher--child data provide a useful setting for studying this question because they contain more than one meaningful form of organization. Children can be grouped according to similarities in their behavioral profiles, but they can also be linked to their teachers. These two forms of structure are related, but they are not necessarily the same. A representation that preserves behavioral similarity among children may not preserve teacher--child correspondence, whereas a representation optimized for teacher--child correspondence may not preserve behavioral phenotypes. Whether different representation learning objectives recover different latent structures from the same psychometric observations has not been examined.

In this study, we address this question using 757 matched teacher--child pairs from the baseline assessment of the Cyprus cohort of the ProW study. We first identify behavioral phenotypes from child SDQ, ASBI, and CBRS item responses using principal component analysis and clustering. We then train a dual-encoder contrastive representation learning framework to learn shared representations that align matched teacher--child pairs directly from item-level questionnaire responses. Finally, we compare these representations with those learned under a multi-task objective that jointly optimizes teacher--child alignment and behavioral prediction. Rather than asking only whether one model performs better than another, we ask a broader question: how does the choice of learning objective influence the latent structure recovered from the same psychometric data?

The study consisted of four stages. First, child questionnaire items were used to identify behavioral phenotypes. Second, teacher and child questionnaires were jointly analyzed using contrastive representation learning to recover teacher--child correspondence. Third, we examined how changing the learning objective affected the recovered latent structure by comparing the original contrastive model with a multi-task extension that jointly optimized teacher--child alignment and behavioral prediction. Figure 1 summarizes the overall analytical workflow.

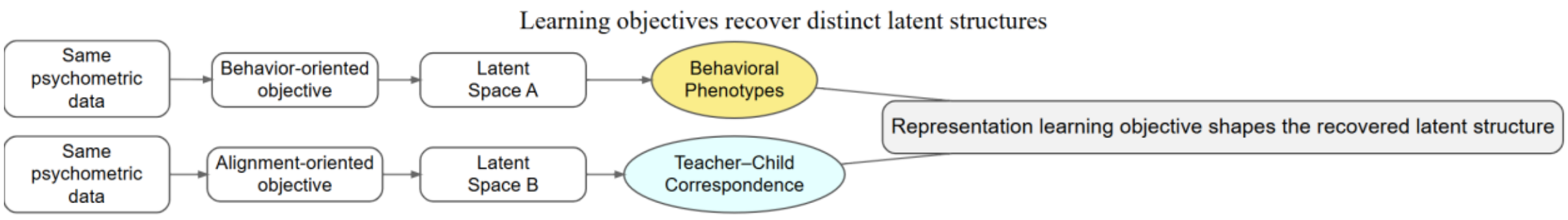

Figure 1. Overview of the proposed dual-encoder contrastive representation learning framework. Teacher and child item-level questionnaire responses are encoded independently into a shared 64-dimensional latent space.

## 2. Method

### 2.1 Dataset and Preprocessing

The analyses were based on Wave 1 (baseline) data from the Cyprus cohort of the ProW project. Sample characteristics are summarized in Table 1-3. The teacher dataset consisted of item-level responses from several questionnaires assessing wellbeing, self-efficacy, job satisfaction, burnout, and professional climate, including the Teacher Subjective Well-Being Questionnaire (TSWQ), PERMA Profiler, Teachers' Sense of Efficacy Scale (TSES), Teacher Social and Emotional Self-Efficacy Scale (TSSES), Job Satisfaction Scale, Maslach Burnout Inventory (MBI), and Professional Climate Scale (PCS). Child data consisted of teacher-reported behavioral assessments derived from the Strengths and Difficulties Questionnaire (SDQ), Adaptive Social Behavior Inventory (ASBI), and Child Behavior Rating Scale (CBRS). After matching teacher and child records, 757 linked teacher--child pairs were available for analysis. In this study, alignment refers to the extent to which a learned representation places matched teacher--child pairs closer together in the latent space than unmatched pairs.

**Table 1: Sample characteristics of the linked teacher–child dataset.**

| Characteristic | Value |
|---|---|
| Teachers | 96 |
| Children | 770 |
| Matched teacher–child pairs | 757 |
| Teachers included in classroom-level analyses | 63 |
| Teacher questionnaire items | 139 |
| Child assessment items | 82 |

**Table 2. Overview of questionnaires and variables**

| Source | Instrument | Construct | Level | Variables used | Analysis stages |
|---|---|---|---|---|---|
| Teacher | TSWQ | Teacher wellbeing | Teacher | Part of 139 teacher items | Stages 2–4 |
| Teacher | PERMA Profiler | Wellbeing | Teacher | Part of 139 teacher items | Stages 2–4 |
| Teacher | TSES | Teaching self-efficacy | Teacher | Part of 139 teacher items | Stages 2–4 |
| Teacher | TSSES | Social-emotional self-efficacy | Teacher | Part of 139 teacher items | Stages 2–4 |
| Teacher | Job Satisfaction Scale | Job satisfaction | Teacher | Part of 139 teacher items | Stages 2–4 |
| Teacher | MBI | Burnout | Teacher | Part of 139 teacher items | Stages 2–4 |
| Teacher | PCS | Professional climate | Teacher | Part of 139 teacher items | Stages 2–4 |
| Child | SDQ | Behavioral difficulties | Child | Part of 82 child items | Stages 1,3,4 |
| Child | ASBI | Adaptive social behavior | Child | Part of 82 child items | Stages 1,3,4 |
| Child | CBRS | Classroom behavior | Child | Part of 82 child items | Stages 1,3,4 |

**Table 3. Data sources and analytical workflow**

| Stage | Input data | Learning objective | Representation analyzed | Primary output | Figures |
|---|---|---|---|---|---|
| Stage 1. Behavioral phenotype discovery | Child questionnaires (82 SDQ, ASBI, and CBRS item responses) | Recover behavioral structure | Child-only latent representation (PCA) | Four behavioral phenotypes | Figs. 2–3 |
| Stage 2. Teacher–classroom behavioral modeling | Teacher questionnaires (139 items) + classroom phenotype composition | Examine associations between teacher characteristics and classroom behavioral composition | Teacher variables and aggregated classroom phenotype measures | Correlation analysis, Random Forest, SHAP | Fig. 4 |
| Stage 3. Contrastive representation learning | Teacher questionnaires (139 items) + child questionnaires (82 items) | Learn teacher–child correspondence | Shared teacher–child latent space | Teacher–child retrieval performance (Top-1, Top-5, Top-10) | Figs. 5–6 |
| Stage 4. Multi-task representation learning | Teacher questionnaires (139 items) + child questionnaires (82 items) + behavioral phenotype labels | Jointly optimize teacher–child correspondence and behavioral prediction | Shared teacher–child latent space | Trade-off between retrieval performance and behavioral organization | Figs. 7–8 |

All analyses were conducted using item-level questionnaire responses rather than aggregated scale scores. Missing values coded as 97, 98, 99, or 999 were treated as missing and imputed using median imputation. Following imputation, all variables were standardized using z-score normalization before representation learning, clustering, and predictive modeling. This preprocessing strategy was intended to preserve information contained in individual questionnaire items while ensuring comparability across measures collected on different scales.

To examine whether behaviorally relevant information was retained within the learned representations, linear probing analyses were conducted using the child embeddings generated by each model. A ridge regression model was trained to predict SDQ outcomes using five-fold cross-validation. Linear probing provided an independent assessment of whether behavioral information could be recovered from the latent representations without additional representation learning.

Throughout this paper, teacher–child correspondence refers to the overall similarity captured between matched teacher and child questionnaire responses. Teacher–child alignment refers specifically to the proximity of matched teacher–child pairs within the learned latent representation, such that matched pairs are embedded closer together than unmatched pairs. Teacher–child retrieval performance is the evaluation metric used to quantify this alignment by measuring whether the correct matched child is retrieved among all candidate children using cosine similarity.

### 2.2 Behavioral Phenotype Modeling

Before learning representations that optimize teacher-child alignment, we first characterized latent behavioral structure within child assessment data. Child-level responses from the Strengths and Difficulties Questionnaire (SDQ), Adaptive Social Behavior Inventory (ASBI), and Child Behavior Rating Scale (CBRS) were combined into an item-level feature matrix consisting of 82 behavioral indicators. To identify low-dimensional behavioral representations, two unsupervised representation learning approaches were evaluated. First, a feed-forward autoencoder was trained using an 82--64--32--16--32--64--82 architecture with Rectified Linear Unit (ReLU) activations and mean squared reconstruction loss. The resulting latent embeddings were visualized using Uniform Manifold Approximation and Projection (UMAP) and clustered using Hierarchical Density-Based Spatial Clustering of Applications with Noise (HDBSCAN).

Second, Principal Component Analysis (PCA) was applied to the standardized behavioral item matrix. The first ten principal components were retained as behavioral embeddings and subsequently clustered using K-means clustering. Candidate clustering solutions were evaluated based on cluster separation and interpretability. The resulting cluster assignments were used to define behavioral phenotypes for downstream analyses. To assess the behavioral validity of the identified phenotypes, cluster membership was compared across SDQ, ASBI, and CBRS outcomes using one-way analysis of variance (ANOVA).

### 2.3 Classroom-Level Behavioral Modeling

To connect teacher characteristics with child behavioral structure, behavioral phenotype assignments were aggregated to the classroom level. For each teacher, the proportion of children belonging to each behavioral phenotype was calculated, yielding a classroom behavioral composition vector. This representation transformed child-level phenotype assignments into classroom-level behavioral profiles and served as the primary link between teacher questionnaire data and child behavioral outcomes.

At the scale level, the strongest association was observed for the Professional Climate Scale (PCS), which was positively associated with the proportion of children belonging to the High Adaptive Functioning phenotype ($r = .33$, $p = .008$). Associations involving wellbeing, burnout, self-efficacy, workplace wellbeing, and job satisfaction were smaller and less consistent. Item-level analyses revealed somewhat stronger relationships, with the largest observed association identified for TSSES 5 ($r = .38$). Overall, the strongest associations were concentrated among indicators of professional climate, wellbeing, and teacher social-emotional self-efficacy.

To evaluate whether teacher characteristics could predict classroom behavioral structure, each classroom was assigned a dominant behavioral phenotype corresponding to the most common child phenotype within that classroom. Random forest classifiers were trained using teacher questionnaire responses as predictors of dominant classroom phenotype membership. Model performance was evaluated using five-fold cross-validation. To improve interpretability, SHapley Additive exPlanations (SHAP) values were computed to quantify the contribution of individual teacher variables to model predictions.

### 2.4 Teacher–Child Contrastive Representation Learning

The contrastive representation learning framework framework is a dual-encoder neural network designed to learn a common latent space for teacher and child questionnaires. During training, responses from matched teacher–child pairs are encouraged to have similar representations, whereas responses from unmatched pairs are pushed farther apart.

Given linked teacher and child questionnaire data, we sought to learn a shared latent representation that captures teacher--child psychometric alignment directly from item-level responses. To this end, we developed a dual-encoder contrastive representation learning framework architecture consisting of separate teacher and child encoders. Teacher inputs comprised 139 questionnaire items spanning wellbeing, workplace wellbeing, self-efficacy, job satisfaction, burnout, and professional climate, whereas child inputs comprised 82 behavioral assessment items derived from the Strengths and Difficulties Questionnaire (SDQ), Adaptive Social Behavior Inventory (ASBI), and Child Behavior Rating Scale (CBRS).

Each questionnaire item was treated as an input token and projected into a 64-dimensional embedding space. Teacher and child responses were processed using independent Transformer encoders, each consisting of two encoder layers with four attention heads per layer. Encoder outputs were aggregated using mean pooling across tokens and subsequently projected into a shared latent space. L2 normalization was applied to obtain the final teacher and child embeddings.

The model was trained using a contrastive learning objective. Within each mini-batch, matched teacher--child pairs were treated as positive examples, whereas all unmatched teacher--child combinations were treated as negative examples. Similarity between embeddings was computed using cosine similarity and optimized using the InfoNCE loss. Training was performed in PyTorch using the Adam optimizer with a learning rate of 0.001, a batch size of 128, and 300 training epochs.

Model performance was evaluated from two complementary perspectives. First, retrieval analyses were used to assess teacher--child alignment within the learned latent space. For each teacher embedding, cosine similarity was computed against all child embeddings, and child embeddings were ranked according to similarity. Retrieval accuracy was quantified using Top-1, Top-5, and Top-10 metrics, corresponding to the proportion of teacher embeddings for which the matched child embedding appeared among the first 1, 5, or 10 retrieved neighbors. Because retrieval performance was the primary outcome of the contrastive learning framework, bootstrap resampling was performed to assess the stability of retrieval estimates. Confidence intervals were calculated from 1,000 bootstrap samples generated by resampling matched teacher--child pairs with replacement.

Cluster quality was assessed using silhouette coefficients, which quantify the extent to which observations are more similar to members of their assigned cluster than to observations belonging to neighboring clusters. Larger silhouette coefficients indicate greater separation between clusters and stronger latent organization. To evaluate whether embedding-derived clusters corresponded to meaningful behavioral differences, cluster assignments were compared across SDQ, ASBI, and CBRS outcomes using one-way analysis of variance. These analyses were intended to determine whether the learned latent space captured behavioral phenotype structure in addition to teacher--child alignment.

All questionnaires used in this study are established instruments with previously reported psychometric validity and reliability. Because the present study focused on representation learning rather than scale development, no additional reliability analyses were conducted.

### 2.5 Multi-Task Teacher--Child Representation Learning

The multi-task extension adds a second learning objective to the original contrastive model. In addition to learning teacher–child alignment, the model is simultaneously trained to predict children's behavioral outcomes from the learned child representations. Thus, the model jointly optimizes two tasks: (1) teacher–child matching and (2) behavioral prediction.

To examine whether explicit behavioral supervision (i.e., adding behavioral prediction as an auxiliary learning task) could improve latent behavioral organization, we developed a multi-task extension of the contrastive representation learning framework. The architecture was identical to the original dual-encoder model, but training was guided by both teacher--child alignment and behavioral outcome prediction objectives. Teacher and child embeddings were first learned using the contrastive learning framework described above. In addition, child embeddings were passed through a prediction head trained to estimate child behavioral outcomes. The multi-task objective jointly optimized contrastive alignment and behavioral prediction during training. The contrastive component encouraged matched teacher--child pairs to occupy nearby locations within the latent space, whereas the supervised component encouraged the learned representations to preserve behaviorally relevant information. Training was performed using the Adam optimizer under the same optimization settings as the original contrastive model. Following training, retrieval analyses, clustering analyses, and linear probing evaluations were repeated to assess the effect of behavioral supervision on both teacher--child alignment and latent behavioral structure.

## 3. Results

### 3.1 Behavioral latent structure recovered from child psychometric data

We first examined whether latent behavioral structure could be recovered from item-level child assessment data. A nonlinear representation learning pipeline consisting of an autoencoder, Uniform Manifold Approximation and Projection (UMAP), and Hierarchical Density-Based Spatial Clustering of Applications with Noise (HDBSCAN) was evaluated as an initial approach. The autoencoder compressed the 82 behavioral indicators into a 16-dimensional latent representation. However, HDBSCAN failed to identify stable clusters and assigned all 770 children to the noise category. Because the resulting latent space did not support meaningful density-based clustering, subsequent analyses focused on a representation learning framework based on Principal Component Analysis (PCA).

PCA revealed substantial low-dimensional structure within the behavioral assessment data. The first principal component explained 37.7% of total variance, and the first ten principal components collectively explained 64.1% of total variance across the 82 behavioral indicators. These ten compo-nents were retained as behavioral embeddings for downstream analyses. K-means clustering applied to the PCA-derived embeddings identified four behavioral phenotypes containing 339, 217, 117, and 97 children, respectively. Figure 2 shows the projection of the first two principal components, with colors indicating the four K-means-derived behavioral phenotypes that are referenced throughout the following analyses.

K-means clustering identified four behavioral phenotypes (Figure 2). In the PCA projection, the purple, green, yellow, and blue clusters correspond to the High Adaptive Functioning, Moderate Adaptive Functioning, Behavioral Vulnerability, and High Behavioral Risk phenotypes, respectively. These phenotype labels are used throughout the remainder of the Results.

As shown in Figure 3, the High Adaptive group exhibited the most favorable behavioral profile, whereas the High Behavioral Risk group showed the least adaptive pattern across measures. To evaluate whether the identified phenotypes captured meaningful behavioral variation, phenotype membership was compared across Strengths and Difficulties Questionnaire (SDQ), Adaptive Social Behavior Inventory (ASBI), and Child Behavior Rating Scale (CBRS) outcomes. Large between-group differences were observed across all three measures. One-way analysis of variance yielded ($F = 161.79$) for SDQ, ($F = 682.54$) for ASBI, and ($F = 834.08$) for CBRS (all $p < 0.001$). The phenotype profiles followed a coherent gradient from highly adaptive functioning to elevated behavioral risk, with progressively higher SDQ scores and lower ASBI and CBRS scores across phenotypes

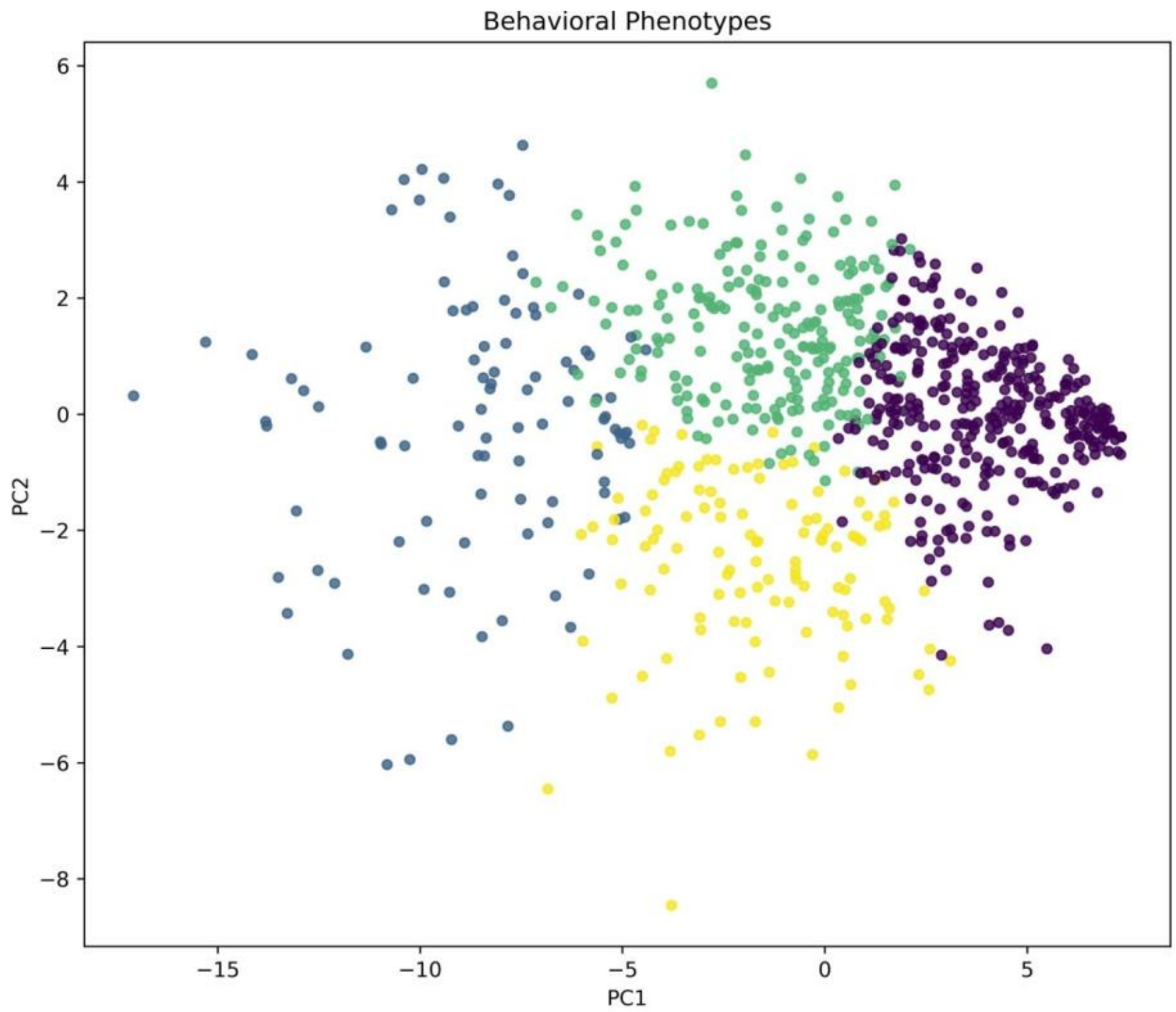

Figure 2. Projection of child assessment embeddings onto the first two principal components (PC1 and PC2). Colors indicate the four behavioral phenotypes identified by K-means clustering: High Adaptive Functioning (purple), Moderate Adaptive Functioning (green), Behavioral Vulnerability (yellow), and High Behavioral Risk (blue).

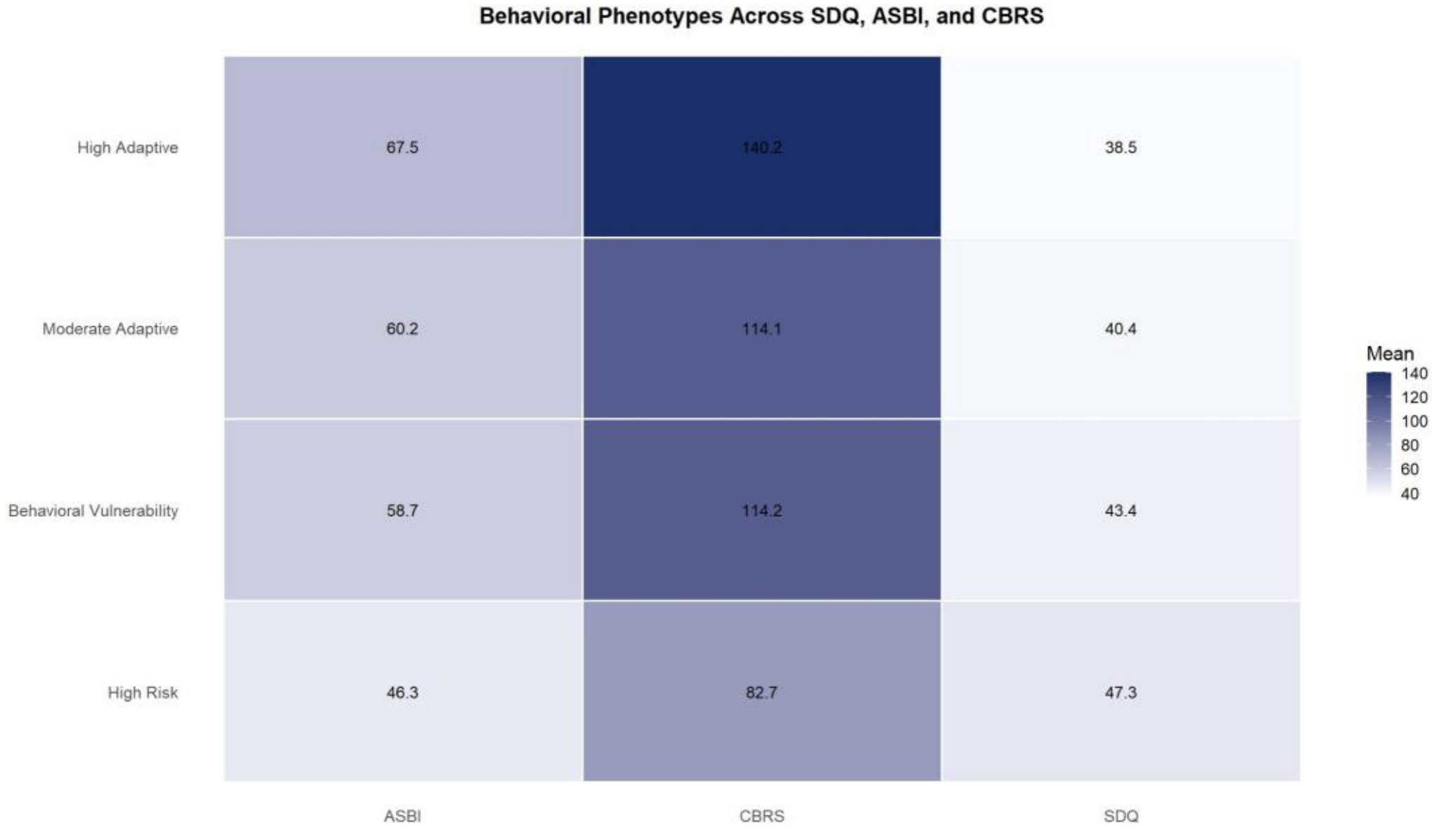

Figure 3. Mean SDQ, ASBI, and CBRS scores for the four behavioral phenotypes shown in Figure 2.

### 3.2 Teacher characteristics explain limited classroom behavioral variation

Behavioral phenotype assignments were aggregated to the classroom level by calculating the pro-portion of children belonging to each phenotype within a teacher's classroom. This produced a four-dimensional representation of classroom behavioral composition that could be linked directly to teacher questionnaire responses. Classroom phenotypes were unevenly distributed. Among the 63 classrooms included in these analyses, 40 were assigned to Cluster 0, whereas only 4, 14, and 5 classrooms were assigned to Clusters 1, 2, and 3, respectively, corresponding to a majority-class baseline accuracy of 63.5

Associations between teacher characteristics and classroom behavioral composition were generally modest (Figure 4). The strongest scale-level association was observed for the Professional Climate Scale (PCS), which was positively associated with the proportion of children belonging to the High Adaptive Functioning phenotype $r = .33$, $p = .008$. Item-level associations were somewhat stronger, with the largest correlation observed for TSSES 5 ($r = .38$). Across analyses, the strongest relationships were concentrated among indicators of professional climate, wellbeing, burnout, and social-emotional self-efficacy.

To evaluate whether teacher characteristics could predict classroom behavioral structure, a random forest classifier was trained using all 139 teacher questionnaire items. Five-fold cross-validation yielded a mean accuracy of 60.5%, which did not exceed the majority-class baseline accuracy of 63.5%. SHAP analyses indicated that the most influential predictors were primarily drawn from measures of wellbeing, professional climate, burnout, and social-emotional self-efficacy, consistent with the correlation analyses. Despite these associations, teacher questionnaire data alone provided limited predictive power for identifying dominant classroom behavioral phenotypes.

Although predictive performance was limited, item-level analyses consistently produced stronger associations than scale-level summaries. Several individual questionnaire items showed stronger relationships with classroom behavioral composition than their corresponding aggregate scales (Supplementary Table 5), suggesting that information relevant to classroom behavioral structure may be partially obscured when questionnaire responses are reduced to scale scores.

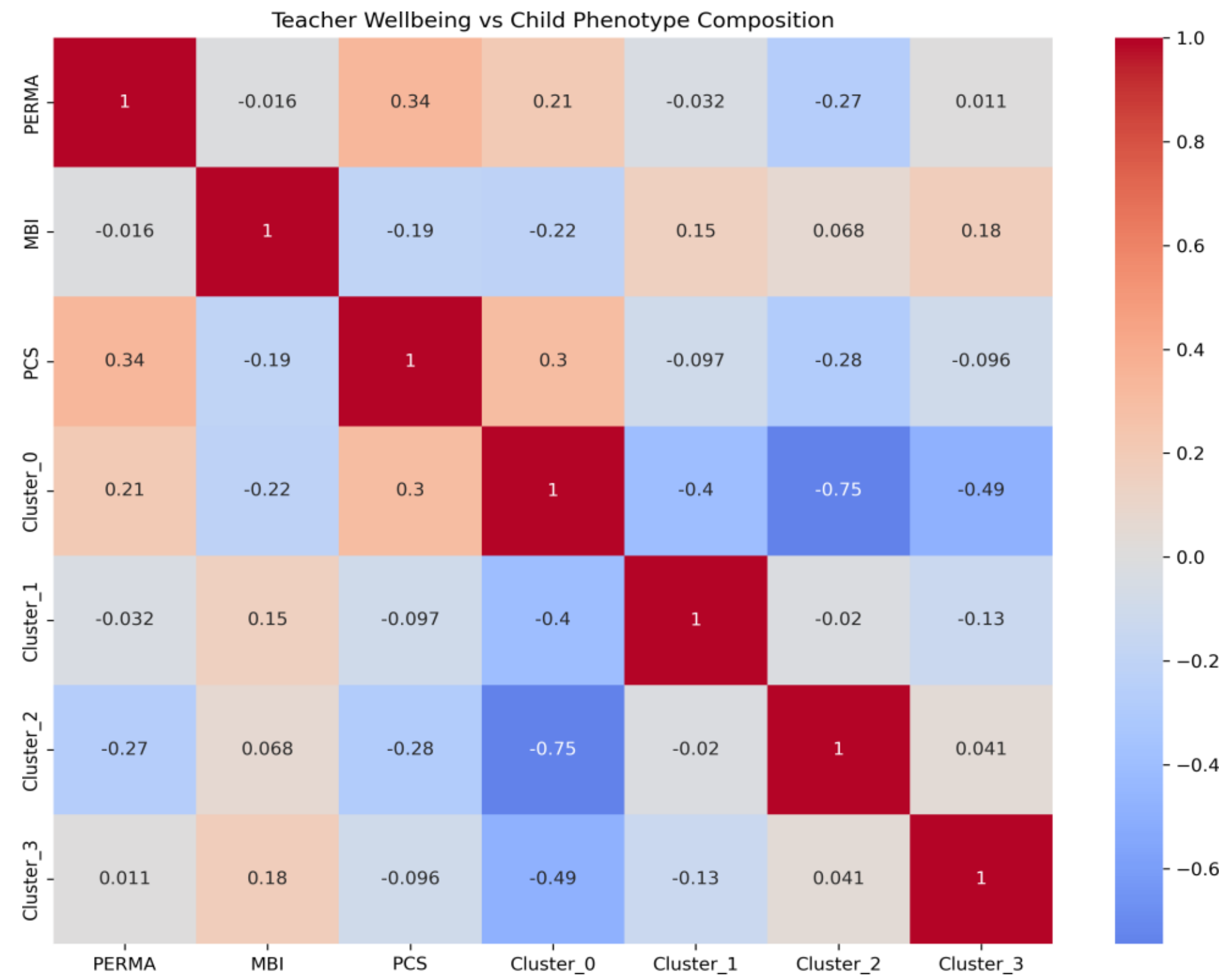

Figure 4: Correlation structure between teacher wellbeing measures and classroom behavioral composition.

### 3.3 Contrastive learning recovers teacher–child correspondence

We next evaluated whether shared teacher–child representations could be learned directly from linked psychometric data. A dual-encoder contrastive representation learning framework was trained using 757 matched teacher–child pairs and evaluated using retrieval-based metrics. During training, the contrastive loss decreased steadily from approximately 29.7 at initialization to 12.8 at the end of training, indicating stable optimization and convergence of the representation learning objective .As shown in Figure 5, the contrastive representation learning framework substantially improved teacher–child retrieval across all Top-k metrics.

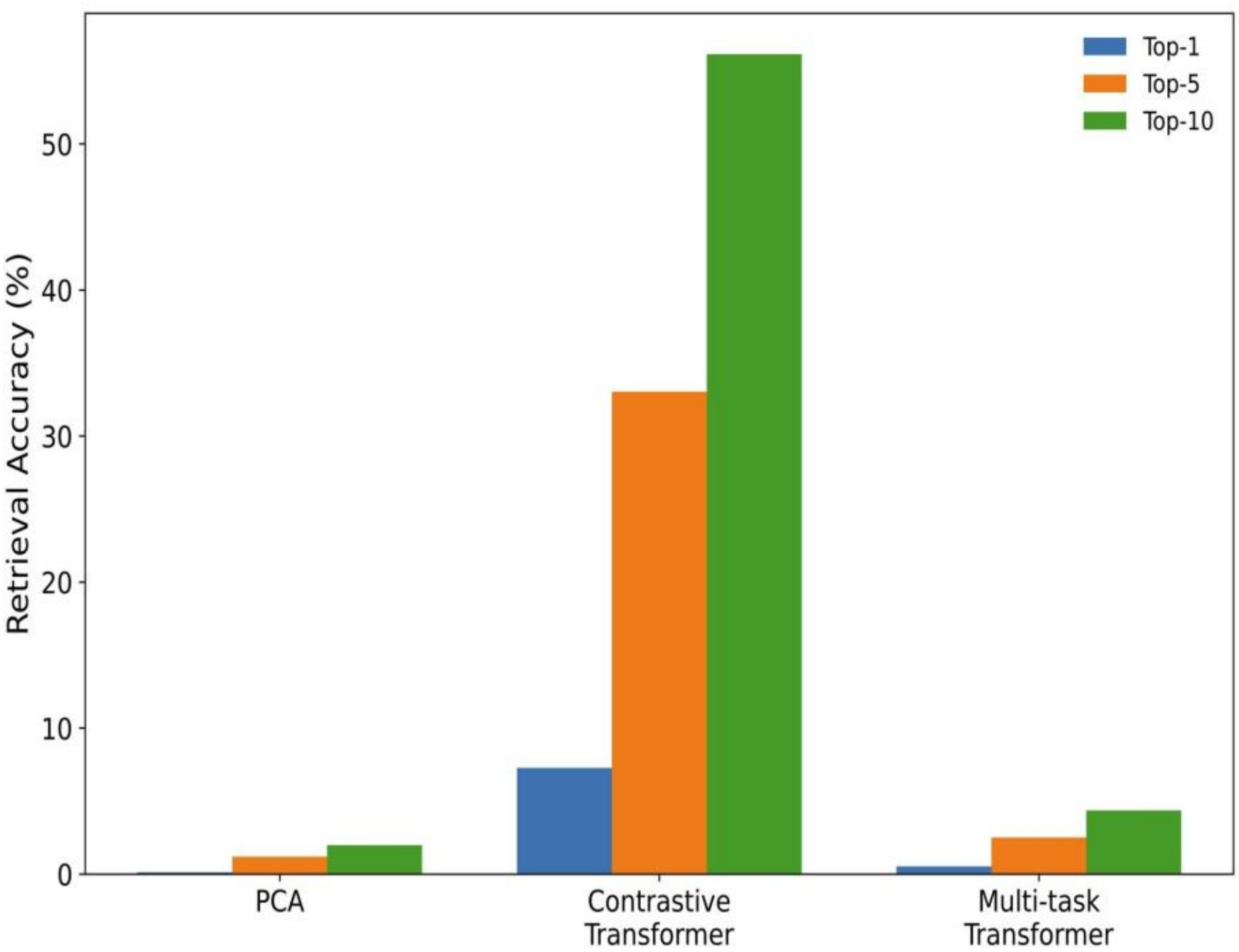

Figure 5. Representations optimized for teacher–child alignment improve retrieval performance. Teacher–child retrieval performance was evaluated using Top-k retrieval accuracy, where k = 1, 5, or 10. Top-1 accuracy represents the percentage of teacher embeddings for which the correctly matched child was ranked first among all 757 candidate children. Top-5 and Top-10 accuracy represent the percentage of teacher embeddings for which the correctly matched child appeared within the five or ten highest-ranked child embeddings, respectively. Rankings were based on cosine similarity in the shared teacher–child latent space.

To assess teacher–child alignment, retrieval analyses were performed in which each teacher embedding was used to rank all child embeddings according to cosine similarity. Retrieval performance was quantified using Top-1, Top-5, and Top-10 accuracy metrics. Top-k retrieval accuracy was used to evaluate teacher–child correspondence, where higher Top-k values indicate that matched teacher–child pairs are positioned closer together in the shared latent space. The contrastive representation learning framework substantially outperformed PCA-based representations across all retrieval measures. PCA embeddings achieved Top-1, Top-5, and Top-10 retrieval accuracies of 0.13%, 1.19%, and 1.98%, respectively. In contrast, the Transformer achieved Top-1, Top-5, and Top-10 accuracies of 7.27%, 33.03%, and 56.14%. The estimated Top-1 retrieval accuracy was 7.27% (95% CI: 6.74–7.53), Top-5 retrieval accuracy was 33.03% (95% CI: 30.38–33.42), and Top-10 retrieval accuracy was 56.14% (95% CI: 51.78–56.54). Bootstrap resampling yielded similar estimates, with narrow 95% confidence intervals across all retrieval metrics, indicating that retrieval performance was stable across resamples (Figure 6).

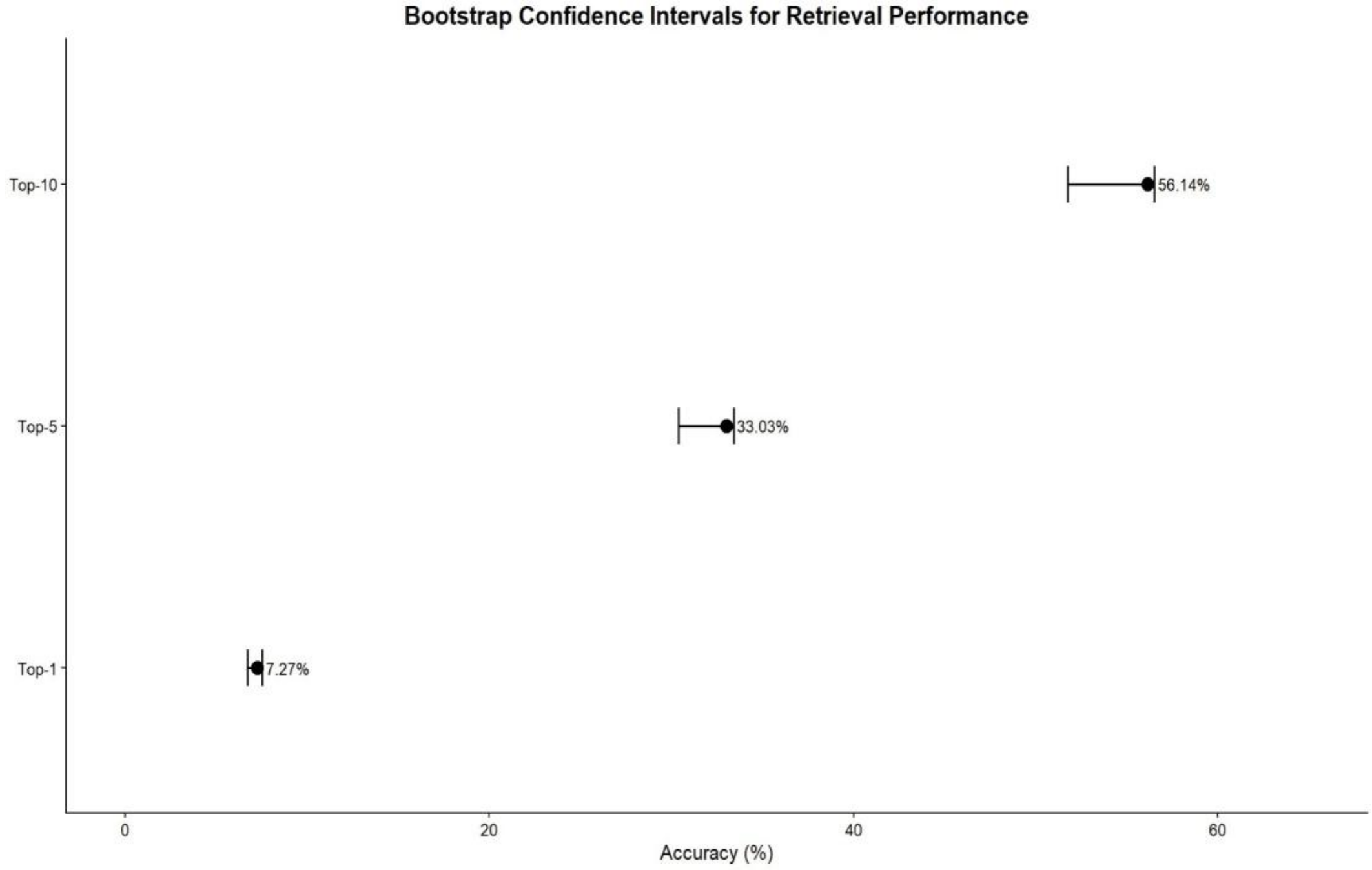

Figure 6: Retrieval estimates are stable across bootstrap resampling, performance with 95% bootstrap confidence intervals. Points represent estimated accuracy, and error bars indicate 95% confidence intervals.

**3.4 Behavioral latent structure is weak under the contrastive objective**

Although the contrastive representation learning framework achieved strong teacher–child retrieval performance, it remained unclear whether behavioral phenotype information was preserved within the learned representations. The resulting four-cluster solution produced relatively balanced cluster sizes (175, 223, 193, and 166 children, respectively), indicating that the weak behavioral separation was not driven by severe class imbalance.

Cluster quality was generally weak across candidate clustering solutions. Silhouette coefficients ranged from 0.076 to 0.085, indicating limited separation between groups in the learned embedding space. Compared with the PCA-derived phenotypes, the resulting clusters showed substantially less behavioral differentiation. One-way analysis of variance revealed no significant differences across Transformer-derived clusters for SDQ outcomes ($F = 1.09$, $p = 0.351$) or ASBI outcomes ($F = 1.53$, $p = 0.206$). A significant difference was observed for CBRS outcomes ($F = 4.53$, $p = 0.004$), although the magnitude of separation remained modest relative to the PCA-derived phenotype structure.

Although behavioral separation was weak, inspection of cluster profiles revealed only small differences across behavioral measures. Mean SDQ scores ranged from 40.55 to 41.29 across clusters, while ASBI scores ranged from 60.65 to 62.20 (see Supplementary Table 5). The largest differences were observed for CBRS outcomes, where Cluster 2 exhibited somewhat higher adaptive functioning scores than the remaining clusters. Overall, the behavioral profiles were substantially less differentiated than those observed for the PCA-derived phenotypes. We further examined whether individual embedding dimensions were associated with behavioral outcomes. Correlations between Transformer embedding dimensions and SDQ, ASBI, and CBRS measures were generally small, with the strongest absolute correlation approximately equal to 0.19. Linear probing analyses produced similar findings, with poor predictive performance for behavioral outcomes. The multi-task objective modestly improved cross-validated prediction of SDQ outcomes, although predictive performance remained poor. These findings suggest that the contrastive representation learning framework primarily learned teacher–child correspondence rather than behavioral phenotype structure.

### 3.5 Changing the learning objective changes the recovered latent structure

To investigate whether behavioral supervision could improve latent behavioral organization, we developed a multi-task Transformer that jointly optimized teacher–child contrastive alignment and behavioral outcome prediction. Training converged successfully, with the total loss decreasing from approximately 5061 at initialization to 50 after 150 epochs. The multi-task objective substantially improved behavioral structure within the learned latent space. Silhouette coefficients increased from approximately 0.08 under the original contrastive objective to values ranging from 0.34 to 0.59. Using a four-cluster solution, the multi-task Transformer identified behavioral groups containing 311, 96, 205, and 145 children, respectively. Behavioral differentiation across clusters also increased. Figure 7 summarizes the resulting trade-off between teacher–child alignment and behavioral organization across the three representation learning objectives.

Figure 7: Representation learning objectives recover different latent structures.

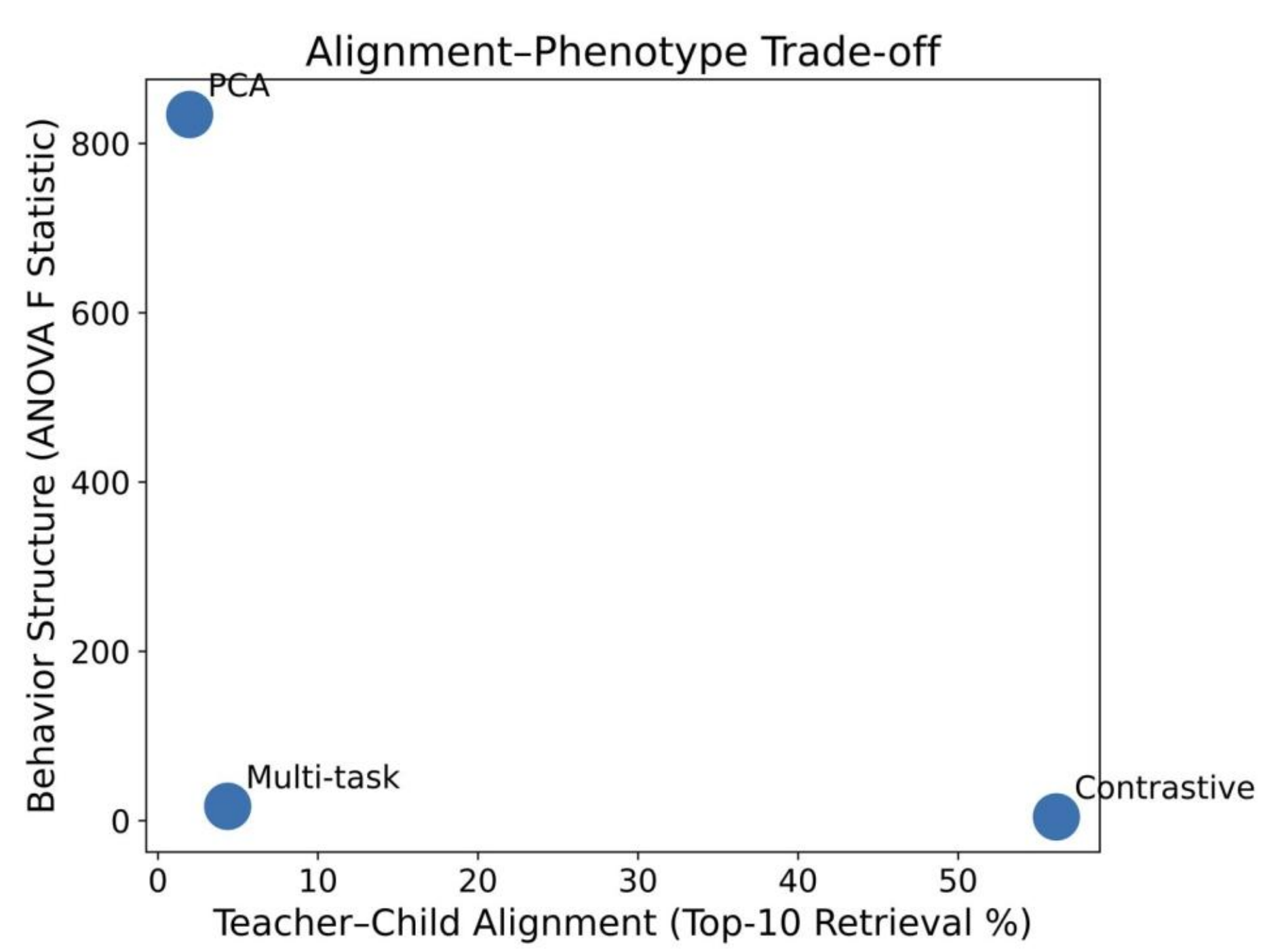

Significant differences were observed for ASBI outcomes ($F = 3.61$, $p = 0.013$) and CBRS out-comes ($F = 16.93$, $p < 0.001$), whereas SDQ differences remained non-significant ($F = 1.48$, $p = 0.217$).The improved behavioral structure was also reflected in the cluster profiles. Cluster 2 exhibited the highest levels of adaptive functioning, with mean ASBI and CBRS scores of 62.99 and 130.78, respectively. In contrast, Cluster 1 exhibited the lowest adaptive profile, with mean ASBI and CBRS scores of 60.29 and 116.60. These differences were consistent with the stronger behavioral separation observed under the multi-task objective.

Although predictive performance remained limited, the multi-task objective consistently im-proved downstream behavioral prediction. Prediction of SDQ outcomes using the learned embed-dings yielded a cross-validated mean $R2 = -0.009$, compared with $R2 = -0.045$ under the original contrastive objective. This pattern suggests that behavioral supervision increased the amount of behaviorally relevant information contained within the learned representations.

However, the improved behavioral organization came at the expense of teacher–child alignment. Retrieval performance declined substantially across all retrieval metrics. Top-10 retrieval accuracy decreased from 56.14% under the original contrastive objective to 4.36% under the multi-task objective. Similar reductions were observed for Top-1 and Top-5 retrieval performance (Figure 8).

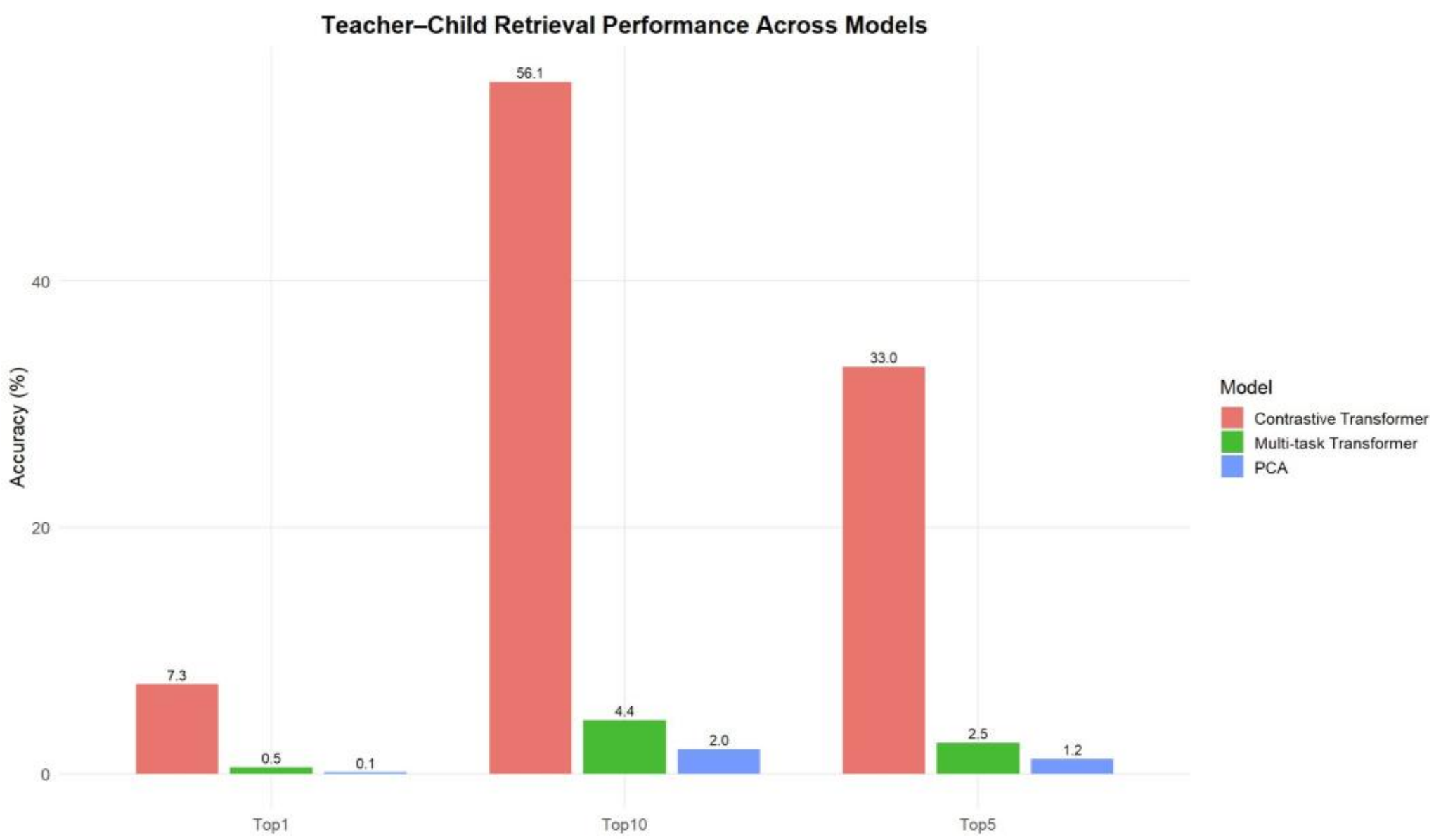

Figure 8: Behavioral supervision improves behavioral organization but reduces alignment performance. Bars represent Top-1, Top-5, and Top-10 retrieval accuracy for PCA, Contrastive representation learning framework, and Multi-task Transformer models.

These findings reveal a tradeoff between teacher–child alignment and behavioral phenotype organization. The original contrastive objective produced the strongest teacher–child correspondence but weak behavioral structure, whereas the multi-task objective improved behavioral organization while substantially reducing retrieval performance. Together, these results suggest that teacher–child alignment and behavioral phenotype discovery represent partially distinct representation learn-ing objectives.

**4 Discussion**

The main finding of this study is that the latent structure recovered from linked psychometric data depended on the representation learning objective. When the model was optimized for teacher--child alignment, it recovered strong teacher--child correspondence but preserved behavioral phenotype structure less effectively. In contrast, PCA recovered clear behavioral phenotypes but provided little information about teacher--child correspondence. Rather than identifying a single underlying representation, these results suggest that different learning objectives emphasize different forms of structure within the same psychometric observations.

Retrieval performance improved substantially under the contrastive objective, although Top-1 accuracy remained modest in absolute terms. This result should be interpreted in the context of the task. Each teacher was compared with 757 candidate children, making exact retrieval inherently difficult. In addition, teachers are associated with multiple children rather than a single child, and several children may reasonably share similar relationships with the same teacher. Classroom environments also introduce variation arising from developmental differences, peer interactions, family influences, and measurement error. Under these conditions, the large improvement over PCA-based retrieval is likely more informative than the absolute Top-1 retrieval rate.

The comparison between PCA and contrastive learning highlights an important difference between representation learning objectives. PCA is designed to preserve the dominant sources of variation within the child behavioral data, whereas contrastive learning explicitly optimizes teacher--child alignment. Consequently, features that distinguish behavioral phenotypes may contribute little to teacher--child matching, while features that improve teacher--child alignment may not strengthen behavioral separation. The multi-task model supports this interpretation. Adding behavioral supervision partially restored behavioral organization but reduced retrieval performance, suggesting that the two objectives compete for representational capacity.

More broadly, these findings suggest that linked teacher--child psychometric data contain more than one meaningful latent structure. Behavioral phenotypes describe similarities among children, whereas teacher--child correspondence describes relationships between teachers and individual children. These structures are related but not identical, and no single representation n recovered both equally well. The present results suggest that the representation recovered from linked psychometric data reflects both the underlying observations and the learning objective used to analyze them. This observation has implications beyond the present application. Representation learning is increasingly used to study complex behavioral data because it can recover structure directly from high-dimensional observations (LeCun et al., 2015; Yarkoni & Westfall, 2017). However, our findings suggest that the recovered structure also depends on the learning objective.

Several limitations should be acknowledged. First, although the original ProW project was conducted across four countries, the present analyses were limited to the baseline assessment of the Cyprus cohort. The generalizability of these findings to other educational settings remains to be established. Second, the analyses relied exclusively on questionnaire responses and did not incorporate observational, environmental, or administrative data that may provide complementary information about teacher--child relationships. Third, only one contrastive architecture and one multi-task formulation were evaluated. Whether the same pattern extends to other representation learning approaches remains an open question. Future work should examine larger and more diverse cohorts, alternative representation learning methods, and multimodal or longitudinal settings to determine whether objective-dependent latent structures generalize across datasets and learning frameworks. These findings may also have broader implications for emerging applications of representation learning in behavioral science, where latent representations are increasingly used to characterize heterogeneous populations (Bzdok & Meyer-Lindenberg, 2018).

## 5 Conclusion

This study shows that linked teacher--child psychometric data contain multiple forms of latent structure. Representations optimized for teacher--child alignment recovered stronger teacher--child correspondence, whereas representations optimized for behavioral organization produced clearer behavioral phenotypes. Neither objective recovered both structures equally well. Rather than identifying a single optimal representation, our findings suggest that the latent structure recovered from psychometric data depends on the representation learning objective. More broadly, these results highlight the importance of matching the learning objective to the scientific question being addressed, rather than assuming that a single representation is optimal for all downstream analyses.

## Declaration of Conflicting Interest

The authors declared no potential conflicts of interest with respect to the research, authorship, and/or publication of this article.

## Funding

The authors received no specific funding for this study.

**Ethical Approval and Consent to Participate**

The data used in this study were drawn from the Wave 1 (baseline) assessment of the Cyprus cohort of the ProW project. All procedures involving human participants were conducted in accordance with applicable ethical standards and the 1964 Helsinki Declaration. The ProW project obtained the necessary ethical approvals from the relevant national and university competent authorities, and participating teachers provided consent to participate. The collected data were pseudonymized prior to analysis.

**Author Contributions**

The authors contributed to the study conception and design, data analysis, interpretation of results, and manuscript preparation. All authors reviewed and approved the final manuscript.

**Declaration of Conflicting Interest**

The authors declare no conflicts of interest.

**Data Availability**

The data used in this study were obtained from the ProW project and are not publicly available because access to the underlying participant-level data is governed by the project's data access and use policies. The data were pseudonymized prior to analysis.

**Code Availability**

Code availability information has been removed for blinded review and will be provided upon acceptance.

**Supplementary Material**

Table 4: Scale-level associations between teacher measures and classroom behavioral composition.

| Scale | Cluster | Spearman (r) | (p) |
|---|---|---|---|
| PCS | Cluster 0 | 0.333 | 0.008 |
| PCS | Cluster 2 | -0.247 | 0.051 |
| MBI | Cluster 0 | -0.242 | 0.056 |
| TSWQ | Cluster 1 | -0.230 | 0.070 |
| PERMA | Cluster 2 | -0.227 | 0.073 |
| TSWQ | Cluster 0 | 0.208 | 0.102 |
| MBI | Cluster 3 | 0.201 | 0.114 |
| MBI | Cluster 1 | 0.190 | 0.135 |
| PERMA | Cluster 0 | 0.189 | 0.139 |
| PCS | Cluster 1 | -0.175 | 0.171 |
| TSSES | Cluster 0 | 0.167 | 0.192 |
| TSSES | Cluster 1 | -0.160 | 0.212 |

Transformer-derived clusters were relatively balanced in size but demonstrated limited behavioral differentiation. Mean SDQ, ASBI, and CBRS scores were similar across clusters, consistent with the weak cluster separation observed in the main analyses. Among the multi-task Transformer clusters, Cluster 2 exhibited the highest levels of adaptive classroom functioning, whereas Cluster

Table 4: Top item-level associations between teacher questionnaire responses and classroom behavioral composition.

| Teacher Item | Cluster | Correlation |
|---|---|---|
| TSSES_5 | Cluster 0 | 0.378 |
| PERMA_20 | Cluster 2 | -0.352 |
| PCS_8 | Cluster 0 | 0.327 |
| PCS_3 | Cluster 0 | 0.319 |
| TSSES_4 | Cluster 0 | 0.318 |
| PERMA_20 | Cluster 0 | 0.317 |
| PCS_21 | Cluster 2 | -0.315 |
| TSSES_20 | Cluster 0 | 0.295 |
| PCS_4 | Cluster 0 | 0.295 |
| TSSES_2 | Cluster 1 | -0.292 |
| PCS_22 | Cluster 2 | -0.288 |
| PCS_11 | Cluster 0 | 0.288 |
| PERMA_10 | Cluster 0 | 0.286 |
| PCS_22 | Cluster 0 | 0.283 |
| TSWQ_2 | Cluster 0 | 0.283 |
| TSSES_1 | Cluster 1 | -0.281 |
| TSSES_28 | Cluster 1 | -0.277 |
| TSSES_18 | Cluster 0 | 0.276 |
| PERMA_1 | Cluster 2 | -0.273 |
| PCS_1 | Cluster 3 | -0.272 |

We exhibited the lowest adaptive profile. These differences were most evident for ASBI and CBRS outcomes. The multi-task Transformer embeddings demonstrated substantially stronger cluster structure than the standard Transformer embeddings. The highest silhouette coefficient was observed at K = 2 (0.595) for the multi-task model, indicating well-defined cluster separation. In contrast, the standard Transformer embeddings exhibited consistently low silhouette coefficients across all cluster solutions, reaching a maximum of only 0.085 at K = 8, suggesting limited intrinsic cluster structure.

Table 5: Top teacher predictors identified by SHAP analysis.

| Feature | Mean SHAP |
|---|---|
| ValuePERMA10 | 0.009228 |
| PERMA_1 | 0.006858 |
| PCS_22 | 0.006490 |
| PERMA_23 | 0.005872 |
| MBI_2 | 0.005269 |
| PERMA_20 | 0.005158 |
| JobSat_4 | 0.005136 |
| PERMA_4 | 0.005007 |
| MBI_1 | 0.004841 |
| TSSES_10 | 0.004712 |
| TSSES_6 | 0.004668 |
| MBI_16 | 0.004590 |
| PERMA_11 | 0.004479 |
| TSSES_19 | 0.003897 |
| PERMA_13 | 0.003850 |
| TSSES_28 | 0.003783 |
| PCS_3 | 0.003778 |
| PERMA_9 | 0.003634 |
| TSSES_5 | 0.003302 |
| PCS_7 | 0.003 |

Table 6: Behavioral characteristics of Transformer-derived clusters.

| Outcome | Cluster 0 | Cluster 1 | Cluster 2 | Cluster 3 |
|---|---|---|---|---|
| N | 175 | 223 | 193 | 166 |
| SDQ | 40.55 | 40.62 | 41.12 | 41.29 |
| ASBI | 61.95 | 60.65 | 62.20 | 61.42 |
| CBRS | 119.07 | 119.38 | 126.21 | 122.71 |

Table 7: Behavioral phenotype characteristics.

| Phenotype | N | SDQ Mean (SD) | ASBI Mean (SD) | CBRS Mean (SD) |
|---|---|---|---|---|
| High Adaptive Functioning | 339 | 38.50 (2.71) | 67.51 (2.67) | 140.17 (10.01) |
| Moderate Adaptive Functioning | 217 | 40.36 (3.73) | 60.18 (4.64) | 114.06 (10.09) |
| Behavioral Vulnerability | 117 | 43.38 (4.46) | 58.66 (5.04) | 114.23 (11.19) |
| High Behavioral Risk | 97 | 47.31 (5.42) | 46.29 (6.02) | 82.72 (13.00) |

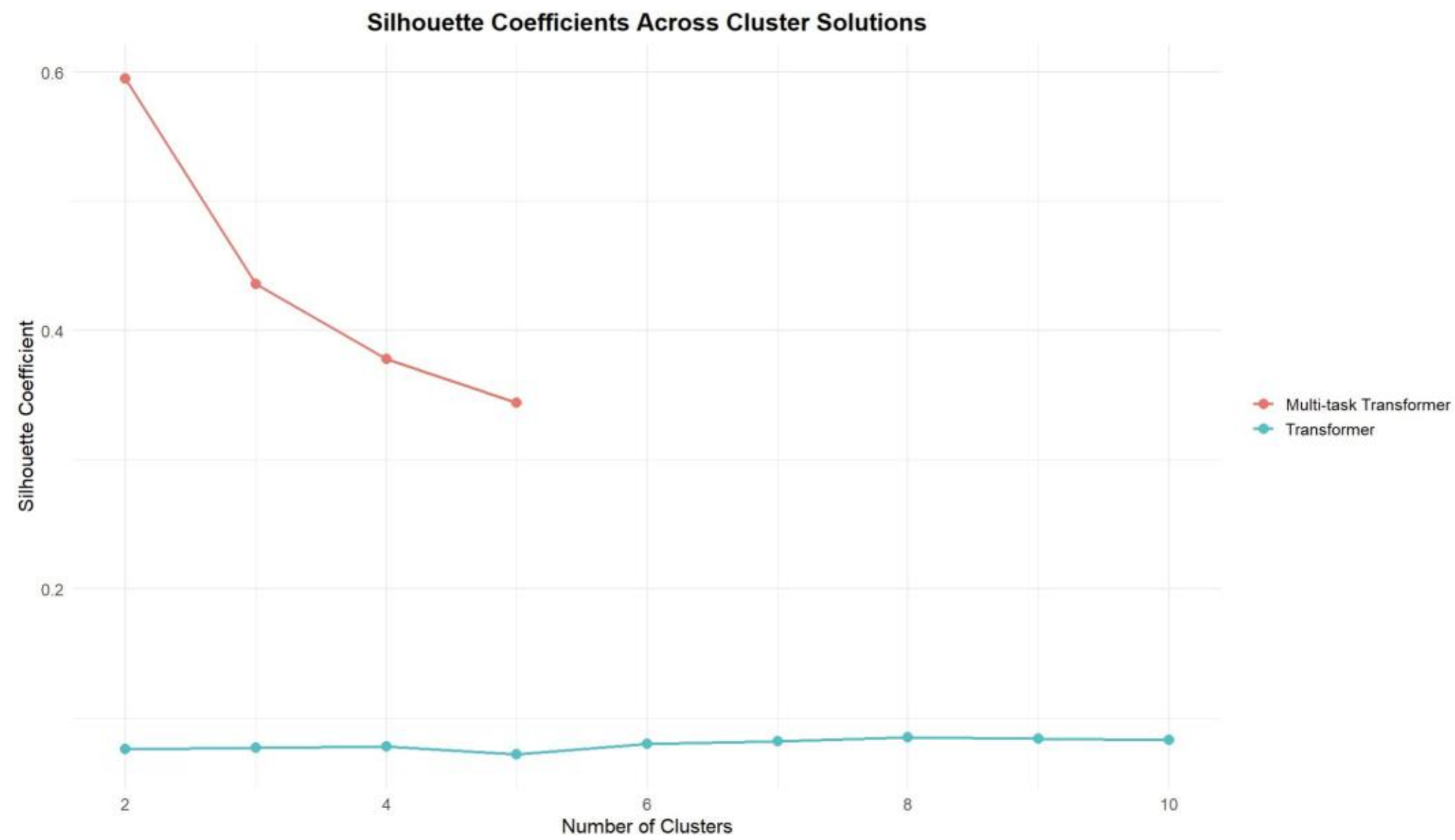

Figure 9: Silhouette coefficients across different cluster solutions for Transformer-derived and multi-task Transformer embeddings. Multi-task embeddings exhibited substantially stronger cluster structure than standard Transformer embeddings, with the optimal solution occurring at K = 2 (silhouette coefficient = 0.595), whereas Transformer embeddings showed weak clustering across all values of K (maximum silhouette coefficient = 0.085 at K = 8).

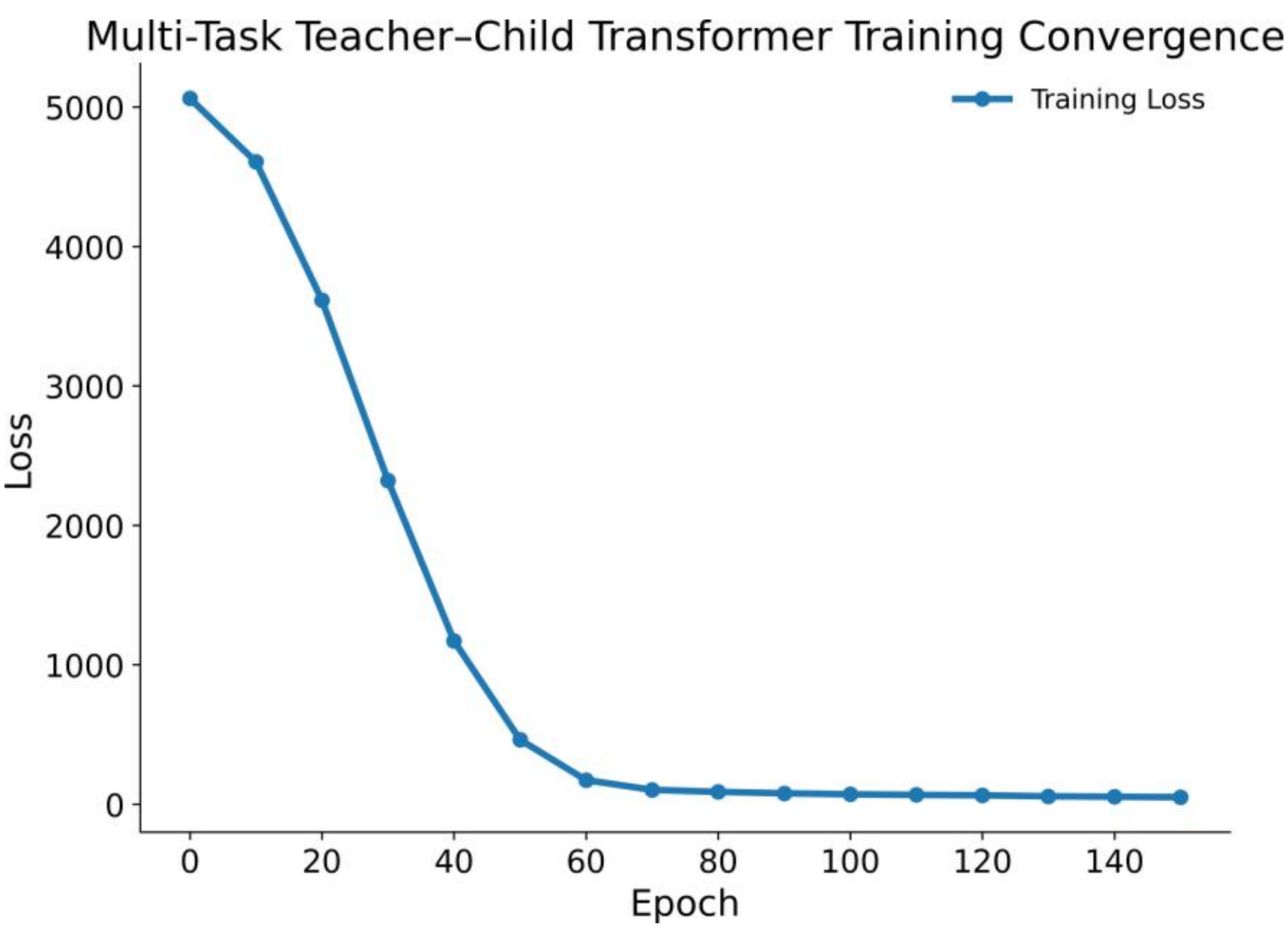

Figure 10: Training convergence of the multi-task teacher–child Transformer. The total training loss decreased rapidly during the early stages of optimization and stabilized after approximately 60 epochs, indicating successful convergence of the learning objective.

Figure 11: UMAP visualization of child embeddings learned by the contrastive representation learning framework. Colors indicate the four PCA/K-means-derived behavioral phenotypes shown in Figure 2. Although the model achieved strong teacher–child retrieval performance, behavioral phenotype separation within the learned latent space remained limited.

Figure 12: UMAP visualization of child embeddings learned by the multi-task Transformer. Behavioral supervision produced improved phenotype separation compared with the original contrastive objective, consistent with increased behavioral organization in the latent space. Colors indicate the four PCA/K-means-derived behavioral phenotype.